PERSPECTIVE

# When evaluation becomes intervention in clinical AI

Anthony Abou Mrad[1], Aviona Conti[1], Jayanth Pratap[1], Ruhika Iyer[2], Joseph Kim[1], King Shi[1], Yann Sakref[3], Luis M. Ahumada[1], Lisa Knowlton[3], Andrew M. Cameron[1], Khyzer Aziz[1], Jeffrey K. Jopling[1]

[1] The Johns Hopkins University, Baltimore, Maryland, USA

[2] Rutgers New Jersey Medical School, Newark, New Jersey, USA

[3] Stanford University, Stanford, California, USA

## Abstract

AI judges, models that grade another system's output against a rubric, extend clinical review to the scale at which clinical AI writes. When teams use judge assessments to select a prompt or train a model, the judge becomes part of the intervention. Selection can then favor errors the judge overlooks. We describe what a judge's report must contain and the independent clinical test each judge-driven change needs.

## Clinical AI and the problem of trust

A clinical model can be trusted only as far as it has been checked against ground truth, the best available way to establish the condition it assesses.[1] For the risk calculators that have long guided practice, that reference was an outcome already present in the data. The American College of Surgeons NSQIP surgical risk calculator was built from 1,414,006 patients at 393 hospitals and predicts outcomes such as death and postoperative complications from 21 preoperative factors.[2] Whether each patient died or had a complication was recorded, so every prediction could be checked against what actually happened. Other tasks lack a recorded outcome, and experts must supply the ground truth. In one study, a model that read retinal photographs for diabetic retinopathy was trained on 128,175 graded images and validated against the majority decision of at least seven board-certified ophthalmologists.[3] Labeling at that scale was expensive, but it gave each image one agreed answer to check the model against.

Language models changed what has to be checked. A model that answers a medical question writes a paragraph, and there is no single correct paragraph to compare it with.[4] Agents are going further by using tools, acting in the health record, and talking directly with patients, as one diagnostic agent did with patients before their primary care visits.[5] For open-ended clinical text, the ground truth is clinical review: clinicians read each output against the source evidence and judge whether it is correct, complete, and safe.

Clinicians cannot review this text at the pace it is produced. One vendor's documentation system alone drafts notes for more than a million encounters a week.[6] To keep pace, the models that write clinical text are now also used to judge it, against clinician-designed rubrics and vetted examples.[7,8] A modest set of clinician judgments can calibrate a judge that then assesses thousands of cases.[9] Judge assessments are also acted on. When one open-weight judge stood in for an expert panel in rating clinical texts, the preferred output changed in 41.7% of pairwise comparisons.[10] Teams use judge assessments deliberately to choose which version goes live or what a model is trained on. They also use them informally: a team that compares versions by their scores and keeps the higher-scoring one is selecting by the judge, whether or not it calls it that. Either way, a judge whose scores select a prompt or train a model has become part of the intervention it was built to measure.

Consider an AI scribe whose note leaves out that the patient was told to seek emergency care if chest pain returns. A judge whose rubric does not assess follow-up instructions could give the note a passing score, and the note would enter the record without the chest-pain instruction. Across eight judge designs, LLM judges distinguished an intact note from the same note with a fact removed little better than chance, though they detected added or altered content well.[11] If the team then keeps the prompt that earns the highest scores, or retrains the scribe on the judge's verdicts, the omission may persist or become more common in future notes. Trusting this judge would take two kinds of evidence. The judge would have to meet prespecified thresholds for detecting clinically important errors. Independent review would also have to measure missed errors among passing notes. A scribe tuned on its scores would have to write notes that are better on what matters clinically, without adding harm.

Existing guidance for clinical AI judges covers how to validate a judge and govern it over its lifecycle.[12] We build on that guidance by separating validation of a judge's measurements from independent testing of the changes its scores drive. We organize the evidence by how a judge's output is used: to observe or inform, to shape or select, or to supply training labels or rewards.

## What an AI judge is, and the roles it plays

### Defining an AI judge

We use the term AI judge for a model that grades another AI system's output against stated criteria and returns an assessment that people or systems then act on.[13–16] In clinical AI, the criteria are usually a clinical rubric. The machine-learning literature calls this LLM-as-a-judge, a term introduced with studies of chatbot evaluation.[7] When the judge uses tools to inspect the steps an agent took as well as its final answer, it is called agent-as-a-judge.[17]

Not every automated check is a judge. Directly verifiable signals, such as dose arithmetic or whether a record reaches a specified state, can be checked against a preset rule. A judge is needed where checking an output requires interpretation. Whether a note captured what mattered in a visit, such as a medication the patient stopped because of side effects, can be decided only by reading the source evidence against clinical criteria. We focus on judges that apply written criteria a clinician can check. Reward models score outputs without written criteria, and when their scores select a version or train a model, the change needs the same independent test.[18]

### How a judge's output is used

Consider again the AI scribe. A judge reviewing its draft notes might score each note on a rubric covering accuracy, completeness, and tone, or rank several drafts. For a single note, the judge's output may route the note to a clinician or trigger a revision before the clinician sees it. Across many notes, the scores describe how the scribe performs. They can also be used to select its prompt, modify its harness (the instructions, tools, and retrieval that surround the model), or train the model itself.[18–24]

A judge's output can serve three overlapping roles (Fig. 1). It can observe or inform, as when it benchmarks a system or provides scores alongside cases a clinician is already reviewing, without changing which cases receive review or their priority. It can shape or select, as when it sets which cases clinicians review and in what order, blocks, ranks, or revises an output, or picks the prompt or configuration that goes live. It can also supply labels or rewards that change the model's learned parameters, or weights.

The roles differ in who or what supplies the evidence for a change. If a clinician reviews the underlying case and decides what to fix, the clinician directs the repair. If the judge's verdict is enough to make the change, the judge drives it, whether it acts on its own or a person carries out its decision. Routing is the exception: even when a clinician directs every repair, a score that decides which cases receive review is a shape-or-select use.

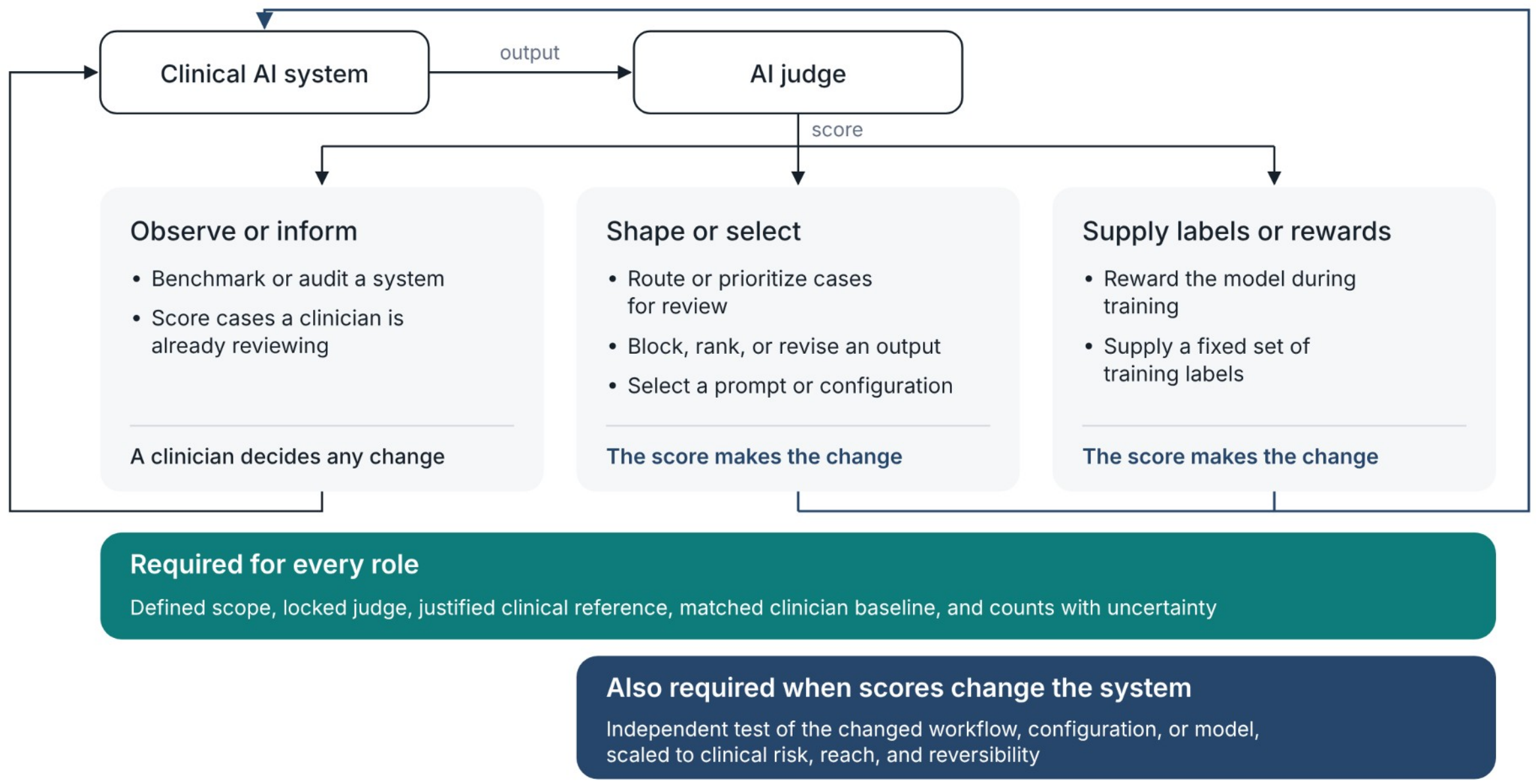

Figure 1 | Roles of an AI judge and the evidence each requires. One judge can hold several roles. Every role requires a report that shows what the judge's scores mean. When the scores change a workflow, configuration, or model, the change must also be tested against clinically relevant performance measures assessed independently of the judge, with evidence scaled to clinical risk, reach, and reversibility.

### What a judge must show

Agreement with clinicians shows how closely a judge follows their judgments on cases like those tested, under the stated rubric. Once the judge's output decides what the system says or which version runs, the claim changes: the resulting change must meet criteria set in advance for clinically relevant performance and harm. Measurement theory treats the consequences of using a score as part of its validity, so that claim needs its own evidence, independent of both the judge and the process that selected the change.[25,26]

## Defining and reporting what a judge measures

### Defining the task and locking the judge

A single score hides which criteria were met.[13] Calibration can help a judge apply its criteria consistently but cannot supply a criterion the rubric omits. Rubrics can narrow during construction toward what can be labeled reliably.[27] In the scribe example, a missing follow-up criterion leaves the dropped chest-pain instruction untested, however well the judge is calibrated. Adding the criterion may not be enough: judges told to check completeness still detected removed facts near chance, and detection partly recovered only when each fact in the transcript was checked against the note.[11] Teams should lock the judge's full configuration before testing (Box 1). In one audit of three commercial scribes, changing only the judge's instruction raised the share of candidate errors it confirmed from 9.3% to 79.0%.[28] A vendor update to the underlying model creates a new version of the judge even when the product name is unchanged.[29,30]

**Box 1 | What a report on an AI judge should contain**

**Scope and configuration**

- Task, patient population, language, care setting, and agent role.
- Locked configuration: model version, full prompts, rubric, example answers, input preparation, decoding settings, tools, and rule for combining item scores.
- Criteria considered and dropped during rubric construction.
- For multi-step encounters: how much earlier context the judge saw, whether it reviewed actions and tool calls as well as the transcript, and the step that introduced each error.
- For panels: each judge's model, version, role, and model family, and how often the judges disagree and make the same errors, measured against independently collected clinician labels.

**Comparison with clinicians**

- Matched-baseline design, blinding, and adjudication method set in advance, with the original ratings retained.
- For each criterion, whether clinician disagreement signals an error or a defensible difference, and how the reference label was set.
- Results by criterion, separately for cases where clinicians agreed and cases where they split, and whether different defensible labels would change the conclusions.
- For yes-or-no criteria: prevalence in clinician ratings, the full two-by-two table with the positive class stated, and counts of indeterminate labels.
- Sensitivity, specificity, predictive values, and an agreement coefficient with confidence intervals, or positive and negative percent agreement when clinician ratings serve only as a comparative method.
- Rules for excluded cases, invalid outputs, and indeterminate labels, and whether verdicts were pooled across criteria.
- Thresholds for the intended use, set in advance.

**Samples**

- What reviewers could see, such as the transcript, the recording, or the record of the agent's actions.
- Results with counts, separately for the matched baseline, the random sample, and enriched high-risk samples.
- Random review of unflagged cases and of agreements, with auditors blinded to the judge's verdict.

**Training labels and monitoring, where they apply**

- Source of each label, rater qualifications, blinding, inter-rater agreement, and adjudication.
- Audit size and frequency, failure prevalence, trigger thresholds, and which changes the monitoring can detect and how quickly.

## Multi-step encounters and panels

In a multi-step encounter, an error introduced at one step can affect care through later actions, and how much earlier context the judge sees changes what it finds.[31] A correct final answer can also follow an unsafe path, so the judge may need the record of actions and tool calls as well as the transcript.[32,33]

A team can rely on one judge or several (Fig. 2). Judges in a separate scoring panel assess outputs independently before their scores are combined, which may reduce errors specific to one judge and reveal disagreement.[34–37] In one evaluation of a pediatric decision-support system, multi-model panels agreed with physicians no more closely than the best single judge.[38] Deliberation lets judges exchange reasons on contested cases but may lead them to converge on an incorrect judgment. Selective routing sends uncertain or high-risk cases to a specialist, a second judge, or a clinician, but it depends on the first judge recognizing which cases need escalation.

Judges from different model families may still share errors,[34,39–41] and shared reference answers that raise agreement[42] could also carry one error across a panel. In the same pediatric evaluation, judges agreed with one another more than with physicians.[38]

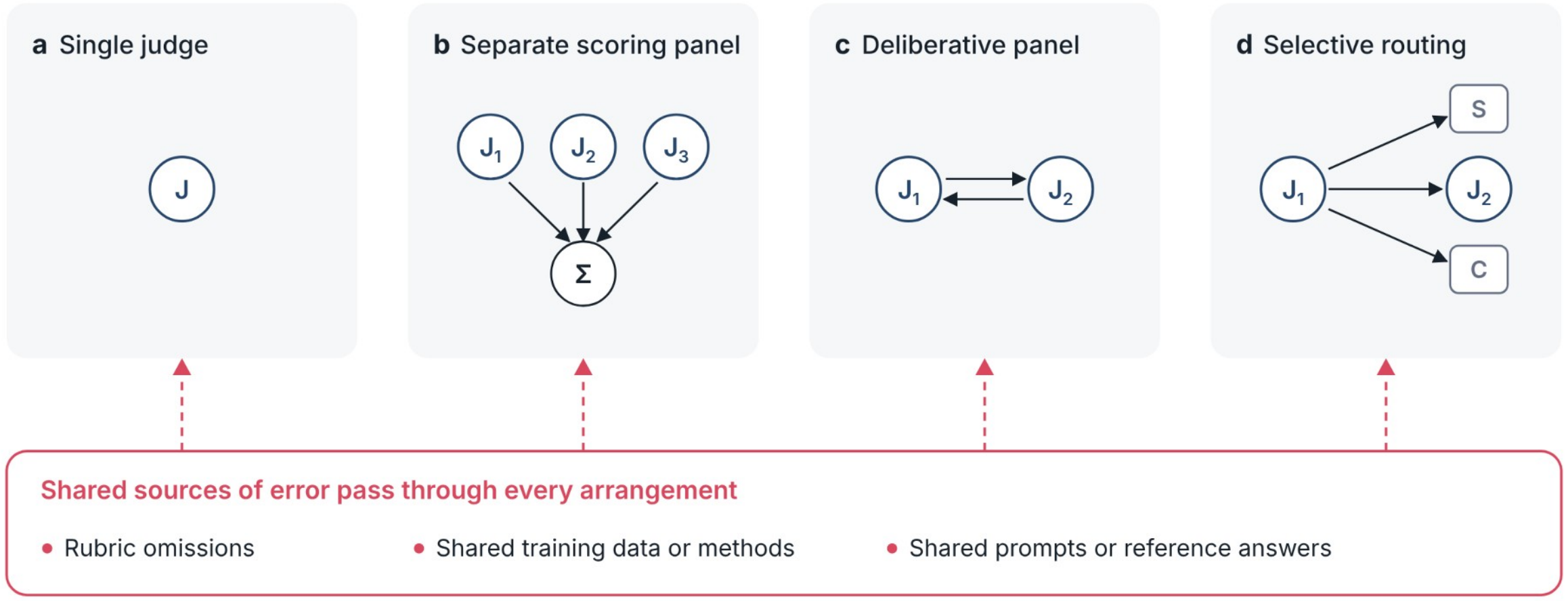

Figure 2 | Judge architectures and shared sources of error. A single judge, separate scoring panel, deliberative panel, and selective routing can be used alone or in combination. Adding judges may reduce errors specific to one judge, but errors the judges share pass through every arrangement. J, $J_1$, $J_2$ and $J_3$ denote judges, and in d, $J_1$ is the first-stage judge and $J_2$ a second judge. Σ, combination of separate scores. S, specialist review. C, clinician review.

## Comparing the judge with clinicians

A fair comparison needs a matched baseline, in which several clinicians independently apply the same rubric to the same in-scope cases. On the first pass, they should be blinded to the judge's score and confidence and to one another's ratings, with the method for resolving disagreements set in advance. This baseline helps distinguish disagreements specific to the judge from variation already present among clinicians.[43] In one audit of 124 urology consultations, three academic urologists applying the same guidelines to the same transcripts reached only slight to fair agreement on which flags from an AI auditor were meaningful, so the auditor's measured accuracy depended on whose judgment served as the reference.[44]

Criteria differ in what clinician disagreement means. For some, such as whether the emergency instruction was preserved, the note either kept it or did not, and disagreement signals an error to resolve through adjudication. For others, such as whether a note is complete enough, clinicians can disagree for defensible reasons, and the disagreement is information to keep. For the first type, the judge is measured against the adjudicated label. For the second, the claim is that the judge falls within the range of clinician judgments, and the report should show where it falls.

For each yes-or-no criterion, the two-by-two table shows what the judge missed, which aggregate agreement can conceal (Fig. 3). In a pediatric decision-support evaluation, judges agreed with physicians on an ordinal safety scale, yet given a pass/fail prompt they passed 89–100% of the answers physicians rated as unsafe.[38] In one study of patient-facing answers, judges separated complete from incomplete responses at near chance, and when judge and clinicians agreed an answer was incomplete, they rarely named the same omission.[16] When adjudicated clinician labels form a reference standard set in advance and independently of the judge, sensitivity is the proportion of reference failures detected, and it should be reported with specificity, predictive values, and an agreement coefficient.[45–48] Otherwise, positive and negative percent agreement replace diagnostic accuracy measures, and predictive values cannot be estimated.[49] A judge suits its intended use only if these estimates meet thresholds set in advance.

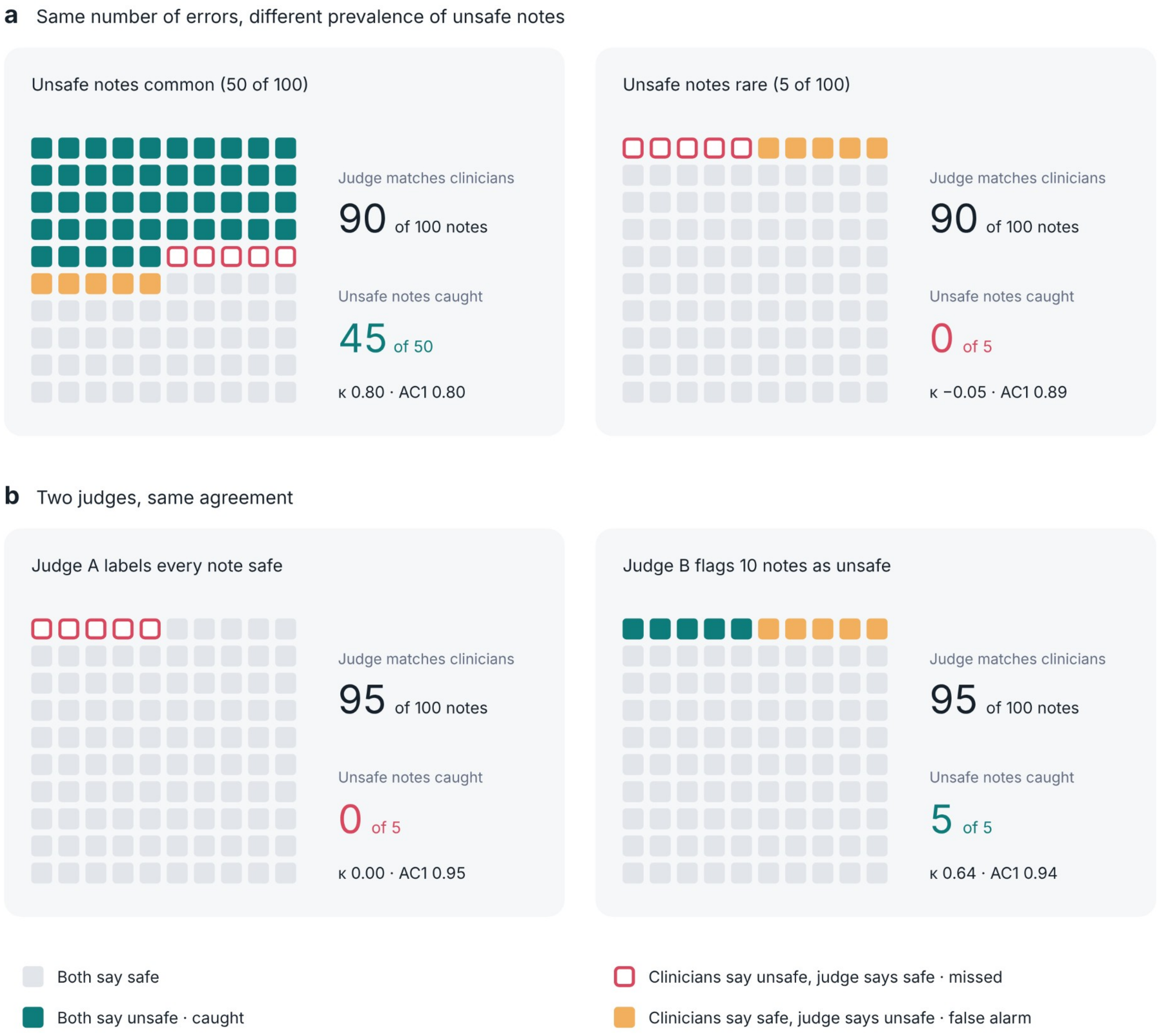

Figure 3 | High agreement can hide missed failures. Each square is one AI-written note, labeled safe or unsafe by clinicians and by a judge for one safety item, whether an emergency-care instruction was preserved. a, In both panels a judge disagrees with clinicians on 10 of 100 notes, with five missed unsafe notes and five false alarms. When unsafe notes are common, sensitivity is 90%, and when they are rare, the same five misses leave it at 0%, while agreement stays at 90%. b, Two judges agree with clinicians on 95 of 100 notes. Judge A misses all five unsafe notes, and judge B catches all five with five false alarms. κ, Cohen's kappa; AC1, Gwet's AC1. The examples are constructed, with clinician labels designated as the reference; sensitivity is relative to those labels. Readers need failure counts and sensitivity to see what a judge missed.[45–48]

### Ongoing review and a complete report

After validation, random review of unflagged cases estimates the proportion of those cases containing missed problems. Targeted high-risk samples should be analyzed separately. Clinicians can adjudicate uncertain cases and panel disagreements, but a judge cannot refer failures it does not recognize. Adjudication therefore cannot replace random review. When disagreement triggers adjudication, a random sample of agreements should also be reviewed, because both sides may have made the same error. Auditors should be blinded to the judge's verdict: in a mammography reader study, incorrect AI suggestions roughly halved experienced radiologists' accuracy.[50] A report is complete when an independent reader can reconstruct the evaluation from it (Box 1).[30]

## Using a judge's output to change the system

### How a system learns to please its judge

When a team uses judge scores to select prompts or train a model, it favors outputs the judge rewards. If the judge rewards an answer despite a clinical error, selection or training may make that error more common, and the judge's score can keep improving while clinically relevant performance plateaus or declines.[51–53] Known judge preferences for longer answers, particular positions, agreement with the user, or their own model family become directions in which the system moves.[39–41,54]

In a medical rubric-training study, rubric-based judges preferred the trained model, whereas rubric-free judges rated its answers lower for factual correctness.[52] In a second clinical-consultation study, a reward that summed rubric items let the model delay the decisive instruction, such as calling emergency services, and recover the points with added advice. Appropriateness on held-out physician criteria fell below the untrained model.[55] Neither study measured harm to patients.

The same thing can happen without retraining the model. A team that tries many versions of a prompt and keeps the one the judge scores highest is also teaching the system to please the judge. Automated tools now run this search at scale and can match the gains of retraining.[56]

### Keeping the test outside the loop

Methods that target evaluator bias, such as judges from different model families[35] or scores corrected for answer length,[41] leave clinical benefit untested, and shared errors can persist.[36] In one study, a prompt that cut judged unsafe overconfidence by 24.7 points also cut correct diagnoses by 2 to 58 points depending on the model, and a judge from another model family put the safety gain at 13.1 points.[57] A judge that guides a change should not supply the only evidence for the changed system's clinical performance, utility, or patient benefit.

The changed system should be compared head to head with the version it would replace, on a sealed set of cases that never informed development. Assessors should be independent of judge development and candidate selection, and blinded to system identity and judge scores. Acceptance or non-inferiority criteria should be set in advance, and harms and subgroup failures assessed separately. Each measure should match the claim, whether output quality, clinical utility, or patient outcomes.

A sealed set stops being independent once its results start guiding revisions, and rotating cases alone does not prevent this. Release decisions should rely on cases withheld from candidate selection or on a prespecified protocol that limits feedback from repeated testing.

Evidence should scale with plausible harm, the patients and decisions reached, clinician oversight, and how quickly a change can be reversed. A narrow, low-risk change may be supported by a blinded offline comparison and monitored rollout. Broad, high-risk changes that are difficult to reverse need prospective evaluation in the intended workflow. Prespecified rollback criteria should reverse deployment when clinical performance falls short, even if judge scores improve.

### Where training material comes from

A judge's output can change a system by updating its weights, through fine-tuning, direct preference optimization (DPO), or reinforcement learning,[18,19] or by rewriting its harness of instructions, examples, and tools.[56] A judge queried during training gives the model repeated chances to exploit errors in the reward. A fixed set of AI-generated labels leaves an auditable record, but its systematic errors may be hard to find. Hybrid schemes can reserve difficult cases for clinicians, provided random audits include AI-labeled cases and the source of each label is recorded.[58]

Several clinical models have been trained on judge scores, with benchmark gains and blinded clinician ratings in two reports.[21–24,59] Among these studies, none reports a prospective comparison isolating judge-generated from clinician-generated training labels using clinical measures independent of the training judge. The closest, ClinAlign, compared physician-revised with automatically drafted rubrics, on benchmarks only.[24]

For high-risk uses involving open-ended clinical judgments, we recommend that judge-trained candidates remain investigational until independent prospective evaluation supports the intended use. When the safety claim is that judge labels can replace clinician labels without material loss, we propose an additional comparison with an otherwise matched model trained on clinician-authored or independently adjudicated labels. The two models should share the base model, cases, label quantity, training methods, compute, and evaluation conditions, and the comparison needs a clinically justified non-inferiority margin. For a purchased model, where no matched counterpart can be built, the comparison is with current care, and the effect of the label source remains unresolved. Adopters should ask vendors where training labels came from.

## Monitoring after release

After deployment, both the clinical system and the judge can change, and monitoring has to tell the two apart.

### Watching the system and the judge

The clinical system changes when its model, instructions, or tools are updated, or when the record templates it reads change. The judge changes when its developer updates the underlying model, when its rubric is revised, or when it is recalibrated on new clinician labels, and a shifting case mix affects both.[60,61] Regression sets can detect the return of known failures, and re-scoring a fixed set of clinician-labeled reference cases can reveal changes in how the judge behaves on those cases.[62] Fixed cases cannot show whether the judge is still reliable on the current case mix, so monitoring should pair them with clinician review of fresh cases. For a judge in the live pathway, the random and high-risk samples described earlier should cover outputs it rejects as well as those it accepts, because harm surveillance cannot rely on agreement with a judge whose rubric may be blind to the harm.

### Detecting change and responding to it

Rising disagreement between the judge and clinicians on random audits may signal drift, a change in the judge's behavior over time. It may also reflect changes in the clinical system or the case mix. For the scribe, clinicians on audit might find more notes missing follow-up instructions while the judge's scores hold steady, a gap that only clinician review would reveal.

Risk-adjusted cumulative-sum (CUSUM) charts track accumulated departures from expected clinical performance,[63] and sequential probability-ratio tests, applied retrospectively to the Bristol and Shipman mortality data, weigh accumulating evidence of a change.[64] We propose applying them to how often clinicians disagree with the judge on audited cases. A threshold crossing, a known failure reappearing, or a report of serious harm should prompt review of the system, judge, rubric, and case mix, proportionate to the risk. Disabling a judge that blocks unsafe outputs or routes urgent cases may remove a safeguard, so the fallback should be verified before release.

## Before a judge’s output changes care

The evidence base is young, as 19 of the 66 references are preprints, and several claims rest mainly on them, including the measured reward hacking in models trained on medical rubrics,[52,53] shared errors among judges from different model families,[34] and methods for separating judge drift from system drift.[30,62] Two references describe one vendor's own practice.[6,8] Our specific evidence requirements are proposals whose feasibility and benefits need testing, and the size of the reported problems should be rechecked as peer-reviewed evidence accumulates.

FDA's guidance on predetermined change control plans (PCCPs) calls for justified reference standards and sequestered test data but sets no validation thresholds for LLM judges.[65] FDA's 2026 discussion paper on generative AI devices extends adjudicator independence to LLM adjudicators, suggests that modified devices could be re-benchmarked, and asks how a machine-based supervisory agent should itself be evaluated.[66] It does not say what evidence a judge-selected change needs. For a judge-selected update covered by an authorized PCCP, we propose that manufacturers document independent clinical evaluation of the change under the plan's Modification Protocol. Journals should require the judge report alongside any claim made for a judge-trained model.

Much of that evidence is still missing. In one review of 49 healthcare studies that used judges, 13 involved no human expert validation, and those that did used a median of three experts.[12] A judge may identify a promising change. Independent clinical evidence should decide whether it reaches patients.

## Data availability

No datasets were generated or analyzed for this Perspective.

## Acknowledgements

This work received no specific funding.

## Use of AI tools

Claude (Anthropic) was used to assist with literature screening, revising the text and checking references against their sources. It was not used to draft the manuscript. The authors reviewed all output and take full responsibility for the content.

## Author contributions

A.A.M. and J.K.J. conceived the Perspective, drafted the manuscript and revised it. A.C., J.P., R.I., J.K., K.S., Y.S. and L.K. contributed to the literature synthesis and revised the manuscript. L.M.A. and K.A. revised the manuscript critically for important intellectual content. A.M.C. and J.K.J. supervised the work. All authors reviewed and approved the final version.

## Competing interests

J.K.J. is an author of a cited benchmark,[33] and K.S. is an author of a cited study.[42] J.K.J. is a co-founder of aprentis, a company developing infrastructure for accountable clinical AI autonomy. All other authors declare no financial or non-financial competing interests.

## Correspondence

Correspondence to Anthony Abou Mrad (aaboumr1@jh.edu).